\pdfoutput=1
\documentclass[letterpaper, 10 pt, conference]{ieeeconf}  

\IEEEoverridecommandlockouts                              

\title{\LARGE \bf
Enhancing Object Detection in Adverse Conditions\newline using Thermal Imaging
}

\author{Kshitij Agrawal$^{1}$ and Anbumani Subramanian$^{2}$}

\author{\authorblockN{Kshitij Agrawal}
\authorblockA{Intel Corporation\\
kshitij.agrawal@intel.com}
\and
\authorblockN{Anbumani Subramanian}
\authorblockA{Intel Corporation\\
anbumani.subramanian@intel.com}}

\usepackage{graphicx}
\usepackage{adjustbox}
\usepackage{cite}

\begin{document}

\maketitle
\thispagestyle{empty}
\pagestyle{empty}

\begin{abstract}

Autonomous driving relies on deriving understanding of objects and scenes through images. These images are often captured by sensors in the visible spectrum. For improved detection capabilities we propose the use of thermal sensors to augment the vision capabilities of an autonomous vehicle. In this paper, we present our investigations on the fusion of visible and thermal spectrum images using a publicly available dataset, and use it to analyze the performance of object recognition on other known driving datasets. We present an comparison of object detection in night time imagery and qualitatively demonstrate that thermal images significantly improve detection accuracies.

\end{abstract}

\section{INTRODUCTION}

 Object detection is one of the primary component for scene understanding in an autonomous vehicle. The detected objects are used to plan the trajectory of a vehicle. Cameras are used to capture images of the environment, which are then input to a myriad of computer vision tasks, including object detection. 

While significant progress has been acheived in using visible spectrum for object detection algorithms, it poses inherent limitations due to the response from cameras in visible spectrum. Some of the shortcomings include low dynamic range, slow exposure adjustment, inefficiencies in high contrast scenes etc, while being subject to weather conditions like fog and rain. Bio inspired vision, like infra-red based thermal vision, could be an effective tool to augment the shortcomings of imagers that operate in the visible spectrum.

Other sensing modalities like LIDAR based systems are sufficient to detect depth in a scene. However, the data may be too coarse to detect objects at further distances and may lack resolution to classify objects. Thermal imagers on the other hand can easily visualize objects that emit infra-red radiation due to their inherent heat. Due to this property, thermal imagers can visualize important participants on the road like people, cars and animals at any time of the day. Augmenting the detection of objects in the thermal spectrum could be a good way to enable robust object detection for safety critical systems like autonomous vehicles.

Object detection methods have progressed significantly over the years from simple contour based methods using support vector machines (SVM) [1-7] to ones using deep classification models \cite{c16,c17,c18,c19,c20} that utilize hierarchal representation of data. Data driven models are the flavor of the day by dominating the detection benchmarks on large scale datasets like PASCAL VOC \cite{c8} and COCO \cite{c9}.

There is a large body of work done for recognizing and localizing objects in the visible spectrum to recognize objects like people [13, 14], vehicles [10] and traffic lights. The features extracted from an image can help identify an object in good lighting and normal weather conditions. However, images obtained using camera systems in low light conditions - night, dusk and dawn, and adverse weather conditions - rain and snow, contain partially illuminated objects, low contrast and low information content. These images are often difficult for object detection algorithms.

The primary contribution of our work is to investigate the nature of object detectors in the thermal spectrum in driving scenarios for autonomous navigation. We utilize the FLIR ADAS [11] dataset that consists of annotated thermal images and time synchronized visible images. Datasets like KAIST [12] exist for similar purpose, however they are limited to annotations of only people.

The next sections are organized as follows: in Section 2, we will cover related research, Section 3, we deal with the datasets, generation of a ground truth for the visible and thermal pairs in the FLIR ADAS dataset and the setup of our experiment. In Section 4 we will present our result and subsequent conclusion in Section 5.

\section{RELATED WORK}


 Object detection consists of recognition and localization of object boundaries within an image. Early work in the computer vision field has focused on building task based classifiers using specific image properties. In some of the earlier approaches a sliding window is used to classify parts of an image based on feature pyramids [15], histogram of oriented gradients (HoG) with a combination of SVM has been used to classify pedestrians [13] and features pools of Haar features [14] have been employed for face detection.

A more generalized form of object detection has evolved over the years due to the advancement in deep learning. The exhaustive search for classification has been replaced by convolutional classifiers. Object detection models have been proposed to work with relative good accuracy on the visible spectrum using models that consist of a) a two stage system  a classifier connected with a region proposal network, RCNN [16], Fast-RCNN [17] and Faster-RCNN [18] b) a single stage network with the classification and localization layers in a cohesive space, like YOLO [19] and SSD [20].

Models trained on large scale datasets are known to perform to quite a good extent. With driving datasets like KITTI [21], Cityscapes [22] the object detection models have been employed to detect pedestrians, cars and bicycles.

Some work has been done in the detection of objects thermal images [23-26], especially focusing on human detection. Since some of the work has been from static camera, the proposals can be generated from background subtraction techniques in the thermal domain [26]. However, most of the work does not deal with investigating the effect of multiple day and night conditions across the thermal and visible spectrum in driving scenarios.

\begin{figure}[tbp]
\centering
\begin{tabular}{cccc}
\includegraphics[height=0.33\linewidth,width=0.44\linewidth]{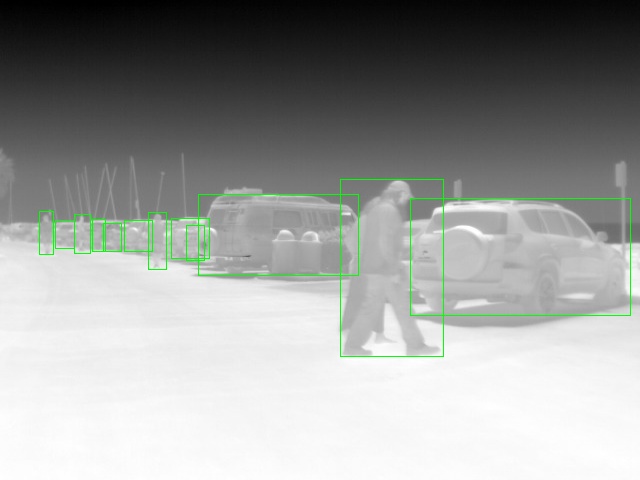}
&
\includegraphics[height=0.33\linewidth,width=0.44\linewidth]{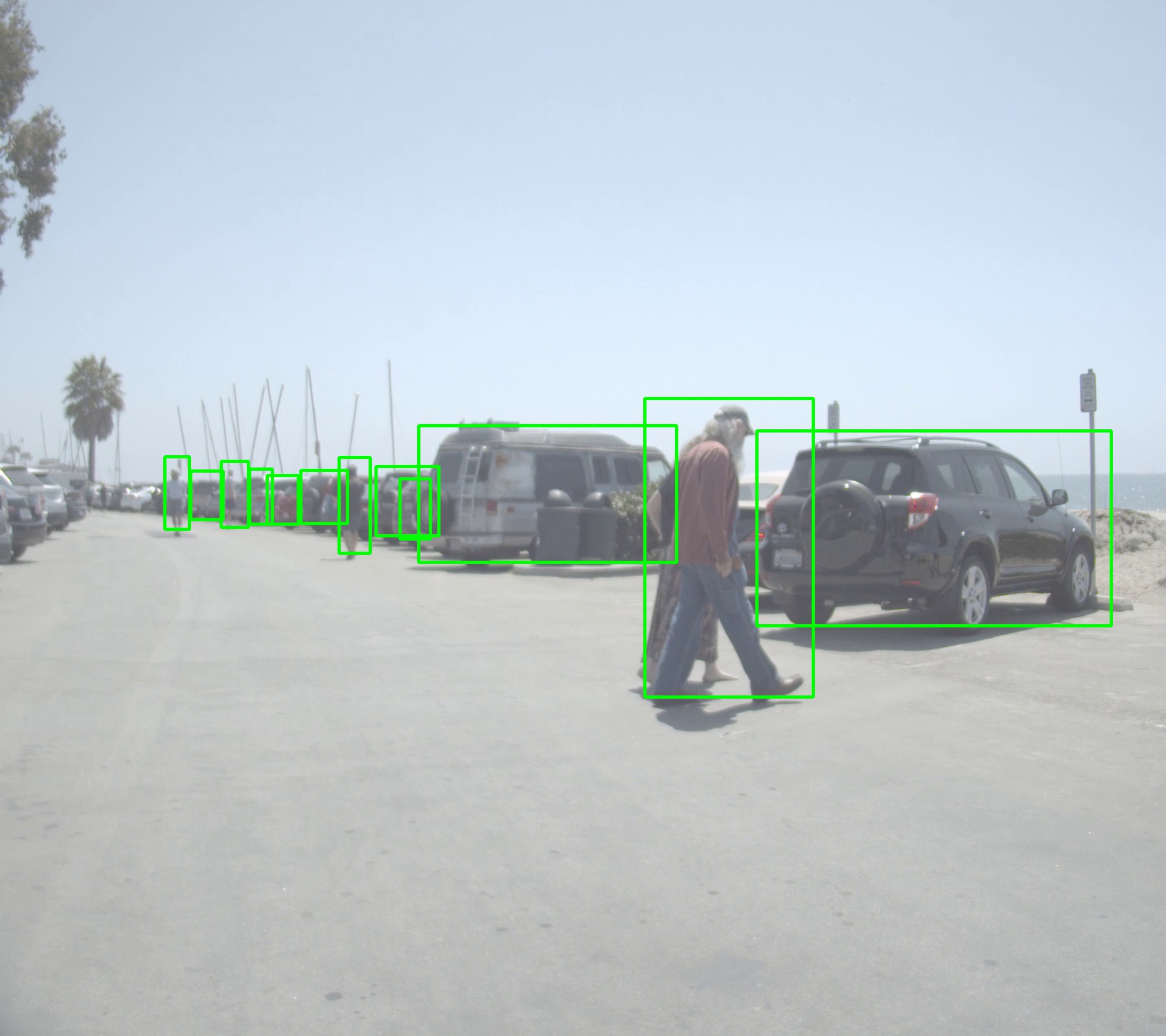} \\

\includegraphics[height=0.33\linewidth,width=0.44\linewidth]{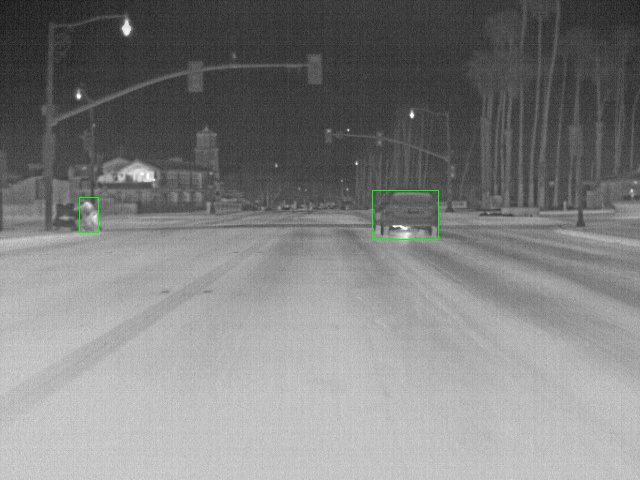}
&
\includegraphics[height=0.33\linewidth,width=0.44\linewidth]{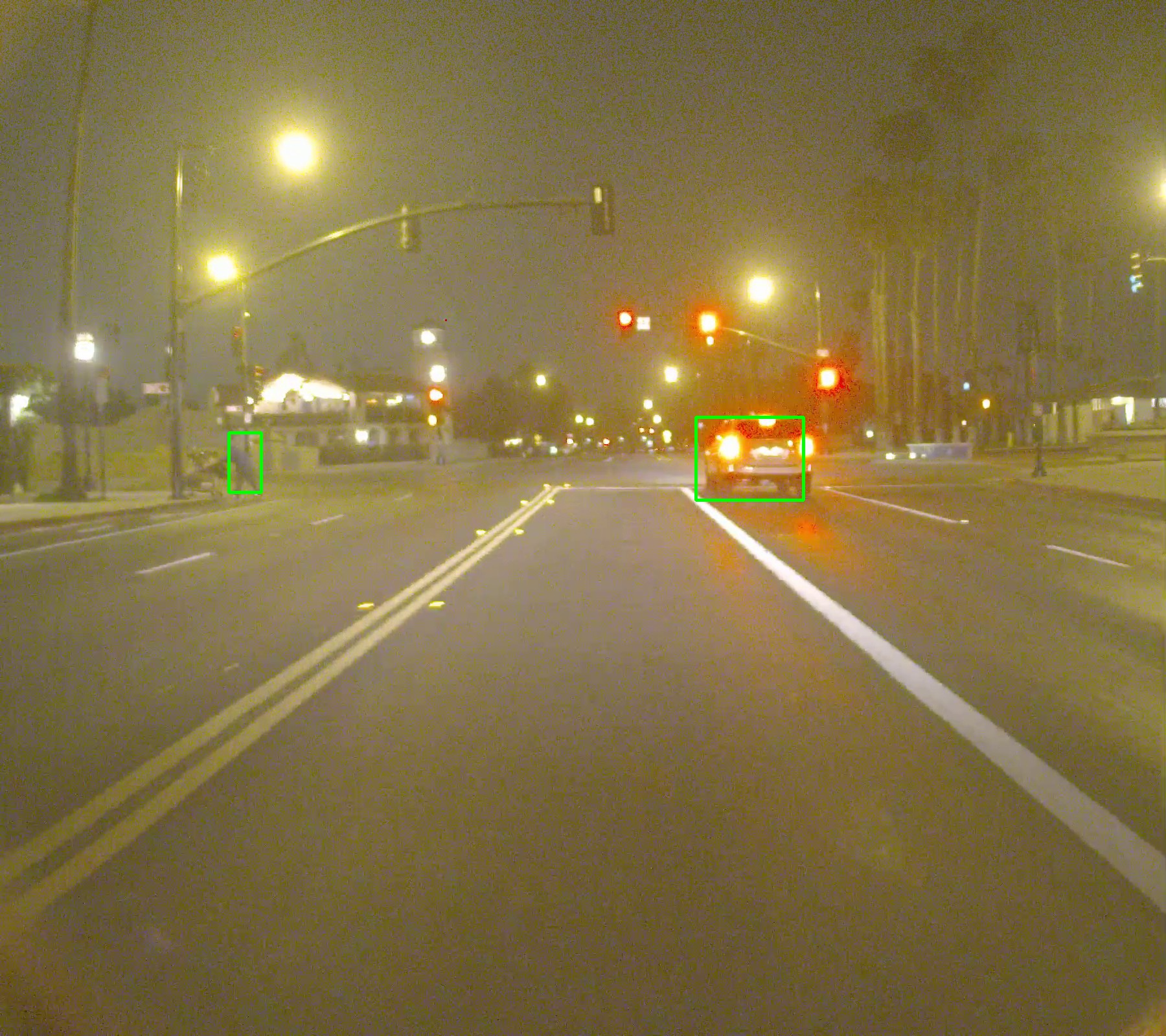} \\
\includegraphics[height=0.33\linewidth,width=0.44\linewidth]{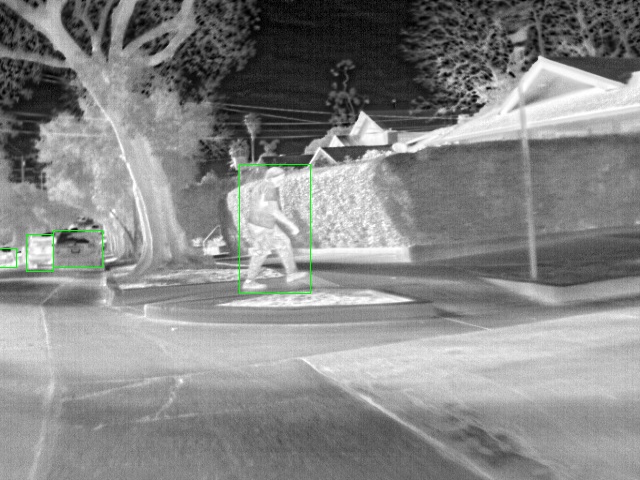}
&
\includegraphics[height=0.33\linewidth,width=0.44\linewidth]{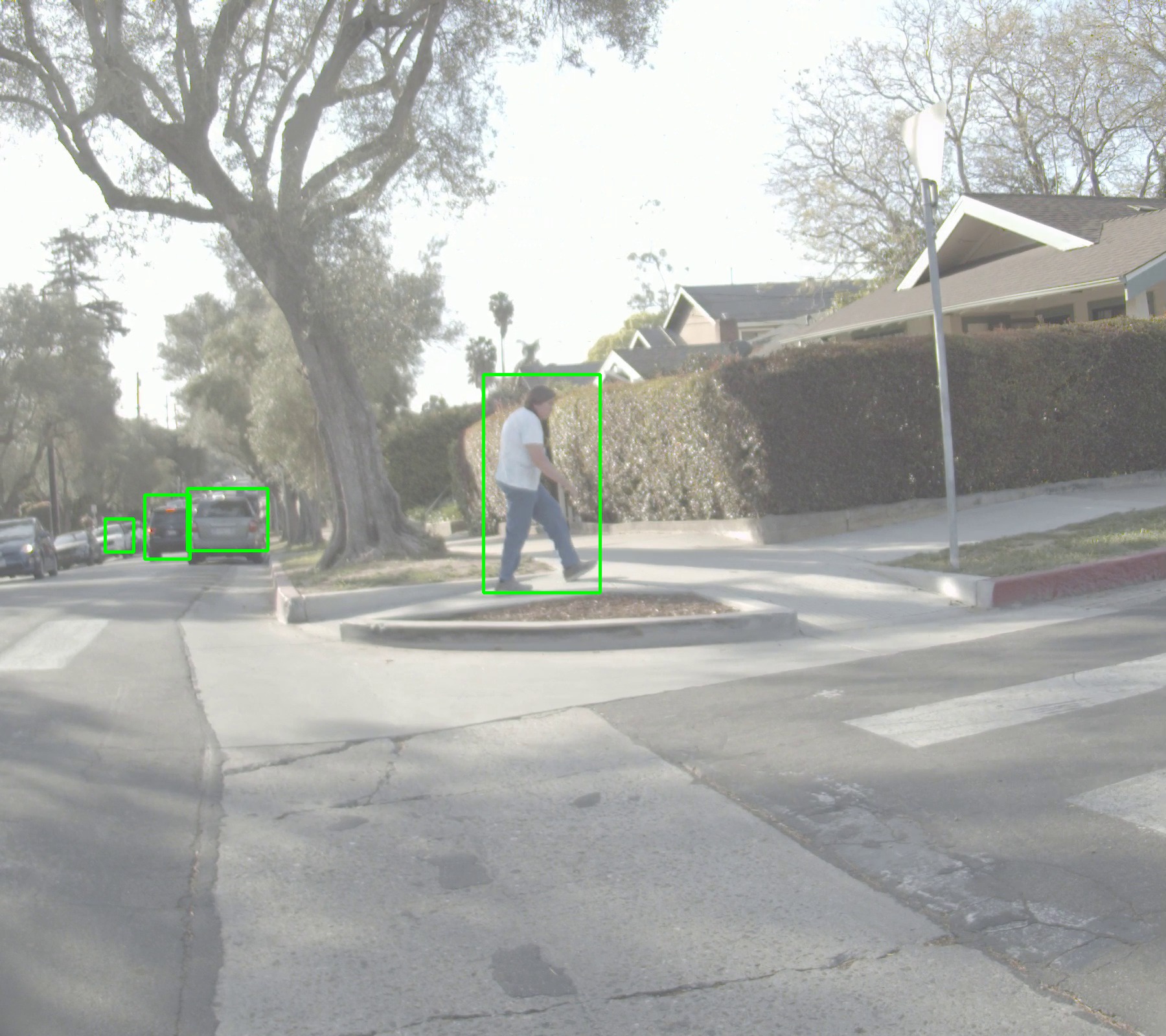} \\

\includegraphics[height=0.33\linewidth,width=0.44\linewidth]{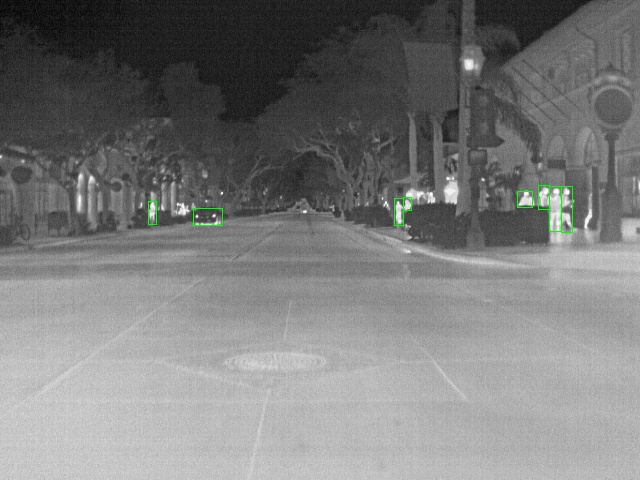}
&
\includegraphics[height=0.33\linewidth,width=0.44\linewidth]{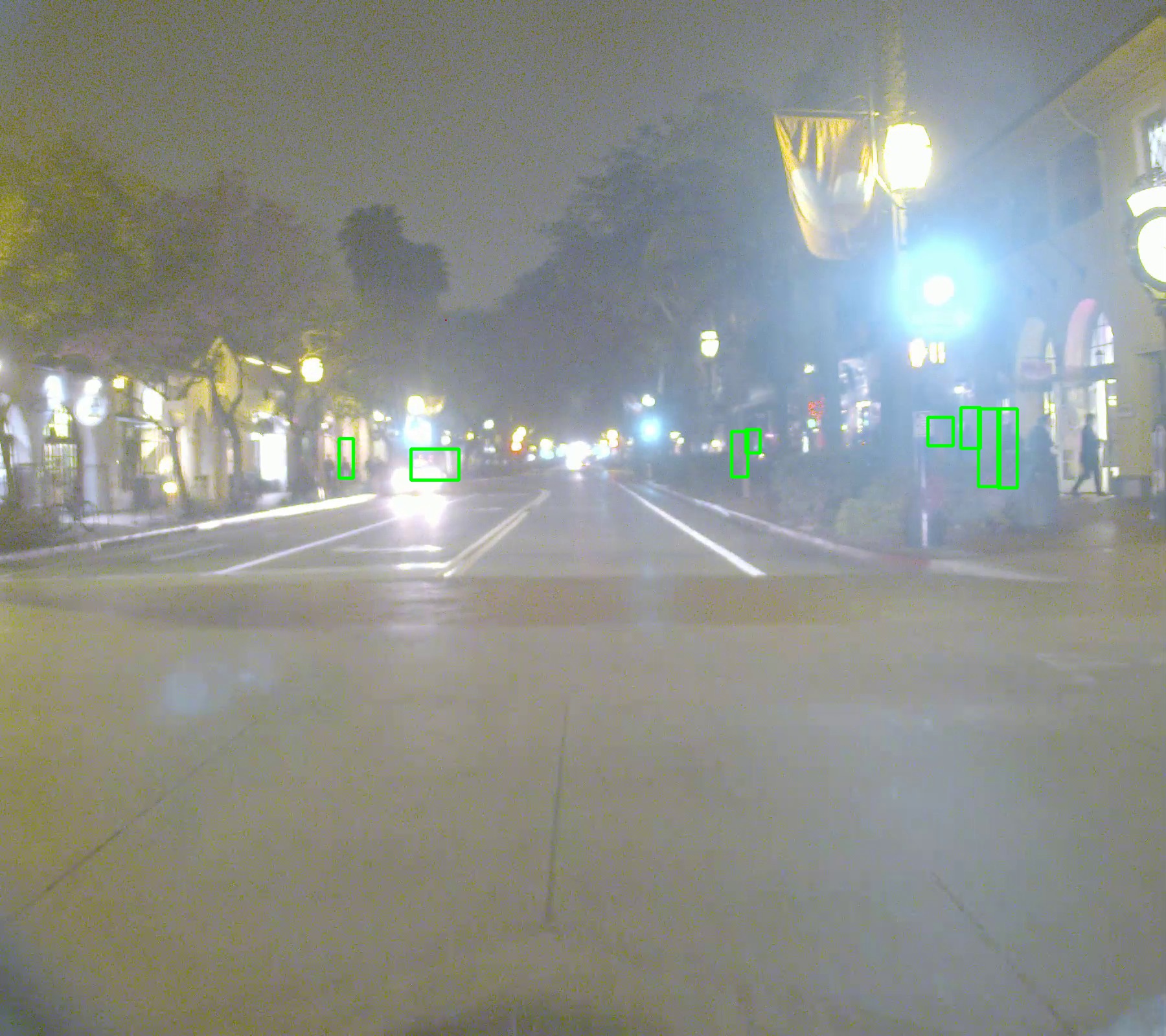}
\end{tabular}
\caption{Annoted and RGB translated pairs from FLIR ADAS dataset}
\label{fig1}
\end{figure}

\section{Datasets}

 In this section we will detail the datasets that we utilize in our study and the process we employed to create a baseline for training the Faster-RCNN model.

\subsection{FLIR ADAS Dataset}

 The FLIR thermal imaging dataset is acquired via a RGB and thermal camera mounted on a vehicle with annotations created for 14,452 thermal images. It primarily is captured in streets and highways in Santa Barbara, California, USA from November to May with clear-sky conditions at both day and night.
Annotations exist for the thermal images based on the COCO annotation scheme. However, no annotations exist for the corresponding visible images.

To analyse the night time performance for object detection it was absolutely essential to have corresponding annotated images in the visible spectrum in the day and night scenarios. 
We build a custom point based correspondence generator and utilized 8 point homography method to generate a correspondence from the thermal to the visible spectrum. Using such methods we are able to translate the annotations to the visible space as well resulting in about 8000 training and 1247 validation images with 42-58 split in night vs day. In the rest of our work we refer to this translated dataset as the FLIR RGB dataset. Fig 1 shows the translation of bounding boxes from the thermal images to the corresponding registered image in the RGB domain.
The input images as part of the FLIR dataset are uncorrected images and slight radial distortions due to the lens can be visualized. The drawback of our technique is that the points closer to the center can be registered, however, the points radially distant from the center do not align well.

\begin{table}[tbhp]
\centering
\caption{Scheme showing mapping of labels}
\label{tab:annotations}
\begin{tabular}{|l|l|l|}
\hline
FLIR    & IDD              & KITTI      \\
\hline
Person  & Person           & Pedestrian \\
        & Rider            & Cyclist    \\
\hline
Car     & Car              & Car        \\
        & Caravan          &            \\
        & Autorickshaw     &            \\
\hline
Bicycle & Bicycle          & -          \\
        & Motorcycle       &            \\
\hline
Dog     & Animal           & -          \\
\hline
-       & Bus              & Truck      \\
        & Trailer          &            \\
        & Truck            &            \\
        & Vehicle fallback &            \\
\hline
\end{tabular}
\end{table}

\subsection{Indian Driving Dataset}

 The India Driving Dataset [27] consists of images taken in driving conditions in city and highway situations primarily during the day. It is unique in the 26 classes that it proposes and the high number of objects in each scene. We pick common traffic participants that also exist in the FLIR dataset and translate them to similar labels. Table 1 shows the translation mechanism. 

\subsection{KITTI}

 The KITTI object detection dataset consist of day time images captured in the urban and highway driving conditions in Karlsruhe, Germany. Again classes corresponding to the FLIR dataset are chosen and translated. A detailed translation can be seen from Table 1. 

%
%
%

\section{EXPERIMENT \& RESULTS}

 The Faster-RCNN implementation from Ren et al [18] was used to train the model on three datasets: FLIR thermal (FLIR\_THM), IDD and KITTI. The Faster-RCNN model used a Resnet-101 for the high level feature extraction and the complete model is initialized from pre-trained COCO weights. The model is trained on each dataset till convergence for about 180,000 iterations. We present the results of each baseline model performance by testing on a validation dataset from the same domain in Table 2. 
 
In the first part of our study the trained model performance is tested on the night time images (653 out of 1247) from the translated FLIR RGB dataset. Table 3, shows that the performance of models trained in the visible spectrum degrades significantly on the night images from the FLIR RGB. We can also see that training on FLIR thermal does not translate well to the visible domain, with a drop of 40\% from the baseline inference on the FLIR thermal dataset. Thus training in the thermal domain does not improve performance in the night time on the same dataset. While training on IDD does retain the highest performance because of better correlation to road scenes from the IDD in day and night conditions. 

\begin{table}[tbhp]
\caption{Average precision per class for dataset combinations tested on night time images from the FLIR RGB translated dataset}
\label{tab:t3}
\begin{adjustbox}{width=\linewidth}
\begin{tabular}{llllll|l}
Train Dataset & Test Dataset & Bicycle & Car    & Dog    & Person & mAP    \\
\hline
FLR\_THM      & FLIR\_RGB    & 0.1312  & 0.571 & 0      & 0.245 & 0.237 \\
IDD           & FLIR\_RGB    & 0.3314  & 0.625 & 0.042 & 0.365 & 0.341 \\
IDD+FLIR\_THM & FLIR\_RGB    & 0.1319  & 0.570 & 0      & 0.260 & 0.240 \\
KITTI         & FLIR\_RGB    & -       & 0      & -      & 0.403 & 0.201 \\
KITTI         & FLR\_THM     & -       & 0      & -      & 0.141 & 0.070 \\
KITTI         & KITTI        & -       & 0.970 & -      & 0.899 & 0.935 \\
\hline
\end{tabular}
\end{adjustbox}
\end{table}

We conduct another evaluation, a performance of domain transfer by introducing the large scale driving dataset into the training. The trained models are tested in the thermal and visible domain for performance gains. We observe a significant drop in performance by testing the IDD and KITTI model on FLIR thermal images - 2.6x drop and 13x drop, respectively. This shows that a model trained in visible domain does not infer well in another domain due to the inherent difference of visual representations. In the case of inference on RGB domain itself we can observe a drop of 1.6x and 6.2x respectively from the baseline performance on the same dataset.

   
\section{CONCLUSIONS}

\begin{table}[bthp]
\caption {Baseline results from training object detection model on the three datasets}
\label{tab:t4}
\begin{adjustbox}{width=\linewidth}
\begin{tabular}{lllllll}
Train Dataset & Test Dataset & Bicycle & Car    & Dog    & Person & mAP    \\
\hline
IDD           & FLR\_RGB     & 0.192  & 0.473 & 0.052 & 0.339 & 0.264 \\
IDD           & FLR\_THM     & 0.126  & 0.265 & 0.099 & 0.160 & 0.163 \\
IDD           & IDD          & 0.569  & 0.617 & 0.070 & 0.448 & 0.426 \\
KITTI         & FLR\_RGB     & -       & 0      & -      & 0.316 & 0.158 \\
KITTI         & FLR\_THM     & -       & 0      & -      & 0.141 & 0.070 \\
KITTI         & KITTI        & -       & 0.970 & -      & 0.899 & 0.935 \\
\hline
\end{tabular}
\end{adjustbox}
\end{table}

From our experiments in Table 2, 3 we can conclude that there is no domain transfer from a model trained in the visible spectrum to inferences in the thermal domain. Thus thermal imagers can prove to be a valuable addition to object detection pipelines, especially for robustness of systems like autonomous vehicles.  
Results in Table 3 also show that few shot training on the Faster-RCNN model from a previously trained model does not perform well across the domains and on new datasets.

\section{Future Work}

Further investigations to evaluate the effect of fusion strategies in the Faster-RCNN network is ongoing. We would also like to compare the effect of multiple fusion strategies with the baseline performance.






\begin{thebibliography}{99}

\bibitem{c1} S. Agarwal and D. Roth. Learning a sparse representation for object detection. In Proc. ECCV, page IV: 113 ff., 2002.
\bibitem{c2} J. Zhang, M. Marszalek, S. Lazebnik, and C. Schmid, "Local features and kernels for classification of texture and object categories: A comprehensive study," in Proceedings of the IEEE Conference on Computer Vision and Pattern Recognition Workshops, 2006.
\bibitem{c3} A. Opelt, A. Pinz, and A. Zisserman, "Learning an alphabet of shape and appearance for multi-class object detection," International Journal of Computer Vision, vol. 80, pp. 16-44, 2008. 
\bibitem{c4} Z. Si, H. Gong, Y. N. Wu, and S. C. Zhu, "Learning mixed templates for object recognition," in Proceedings of the IEEE Conference on Computer Vision and Pattern Recognition, 2009.
\bibitem{c5} J. Shotton, "Contour and texture for visual recognition of object categories," Doctoral of Philosphy, Queen's College, University of Cambridge, Cambridge, 2007.
\bibitem{c6} J. Shotton, A. Blake, and R. Cipolla, "Multiscale categorical object recognition using contour fragments," IEEE Transactions on Pattern Analysis and Machine Intelligence, vol. 30, pp. 1270-1281, 2008.
\bibitem{c7} K. Schindler and D. Suter, "Object detection by global contour shape," Pattern Recognition, vol. 41, 2008.
\bibitem{c8} M. Everingham, L. Van Gool, C. K. I. Williams, J. Winn, and A. Zisserman, "The PASCAL Visual Object Classes Challenge 2007 (VOC2007) Results," http://www.pascalnetwork.org/challenges/VOC/voc2007/workshop/index.html
\bibitem{c9} T.-Y. Lin et al., "Microsoft COCO: Common objects in context," in Proc. Eur. Conf. Computer Vision, Springer, 2014, pp. 740755
\bibitem{c10} S. Gupte, O. Masoud, R. F. K. Martin, and N. P. Papanikolopoulos. "Detection and classification of vehicles", IEEE Transactions on Intelligent Transportation Systems, 3(1):3747, Mar. 2002.
\bibitem{c11} https://www.flir.in/oem/adas/adas-dataset-form/
\bibitem{c12} S. Hwang, J. Park, N. Kim, Y. Choi, and I. S. Kweon, "Multispectral pedestrian detection: Benchmark dataset and baseline," in Proc. IEEE Conf. Computer Vision Pattern Recognition (CVPR), Jun. 2015.
\bibitem{c13} N. Dalal and B. Triggs. "Histograms of Oriented Gradients for Human Detection," In CVPR, 2005
\bibitem{c14} P. Viola and M. Jones,  "Robust Real-time Face Detection", IJCV, 57(2), 2004
\bibitem{c15} P. Dollar, R. Appel, S. Belongie, and P. Perona, "Fast Feature Pyramids for Object Detection," TPAMI, 36(8), 2014
\bibitem{c16} R. Girshick, J. Donahue, T. Darrell, and J. Malik, "Rich feature hierarchies for accurate object detection and semantic segmentation," in Proc. IEEE Conf. Computer Vision and Pattern Recognition, 2014 
\bibitem{c17} R. Girshick, "Fast R-CNN," in Proc. IEEE Conf. Computer Vision, 2015, pp. 14401448 
\bibitem{c18} S. Ren, K. He, R. Girshick, and J. Sun, "Faster R-CNN: Towards realtime object detection with region proposal networks," in Advances in Neural Information Processing Systems, 2015, pp. 9199.J. 
\bibitem{c19} Redmon, S. Divvala, R. Girshick, and A. Farhadi, "You only look once: Unified, real-time object detection," in Proc. IEEE Conf. Computer Vision and Pattern Recognition (CVPR), 2016, pp. 779788.
\bibitem{c20} W.Liu, D.Angeuelov, D.Erhan, C.Szegedy, S.Reed, C-Y.Fu and A.C.Berg, "SSD: Single shot multibox detector," in Proc. Eur. Conf. Computer Vision. Springer, 2016, pp. 2137
\bibitem{c21} A. Geiger, P. Lenz, C. Stiller, and R. Urtasun, "Vision meets robotics: The KITTI dataset," Int. J. Robot. Res., vol. 32, no. 11, pp. 12311237, 2013.
\bibitem{c22} M. Cordts et al., "The cityscapes dataset for semantic urban scene understanding," in Proc. IEEE Conf. Computer Vision Pattern Recognition (CVPR), Jun. 2015, pp. 32133223.
\bibitem{c23}  J. Ge, Y. Luo, and G. Tei. Real-Time Pedestrian Detection and Tracking at Nighttime for Driver-Assistance Systems. IEEE Transactions on ITS, 10(2), 2009.
\bibitem{c24} Y. Lee, Y. Chan, L. Fu, and P. Hsiao. Near-Infrared-Based Nighttime Pedestrian Detection Using Grouped Part Models. IEEE Transactions on ITS, 16(4), 2015.
\bibitem{c25} R. Miezianko and D. Pokrajac. People Detection in Low Resolution Infrared Videos. In CVPR Workshops, 2008
\bibitem{c26} M. Teutsch, T. Muller, M. Huber, and J. Beyerer. Low resolution person detection with a moving thermal infrared camera by hot spot classification. In CVPR Workshops, 2014.
\bibitem{c27} G. Varma, A. Subramanian, A. Namboodiri, M. Chandraker and C V Jawahar, "IDD: A Dataset for Exploring Problems of Autonomous Navigation in Unconstrained Environments", in IEEE Winter Conf. on Applications of Computer Vision (WACV 2019)

\end{thebibliography}
\end{document}